\documentclass{llncs}
\usepackage{makeidx}  % allows for indexgeneration

\usepackage{xcolor}
\usepackage{graphicx}
\usepackage{multicol}
\usepackage{amsmath}
\usepackage{array}
 \usepackage{subcaption}

\usepackage{mathtools}
\DeclarePairedDelimiter\floor{\lfloor}{\rfloor}

\newcolumntype{C}[1]{>{\centering\arraybackslash}m{#1}}

\begin{document}
\frontmatter          % for the preliminaries
\pagestyle{headings}  % switches on printing of running heads
\hyphenation{Tra-ver-sa-ble}
\mainmatter              % start of the contributions
\title{Pushing the Limits of Learning-based Traversability Analysis\\for Autonomous Driving on CPU}
%

% ALTRA IDEA PER TITOLO
% Pushing the limits of traversability analysis for autonomous vehicles: a lightweight real-time implementation on CPU.

%\titlerunning{Hamiltonian Mechanics}  % abbreviated title (for running head)
\author{Daniel Fusaro$^{1}$ \and Emilio Olivastri$^{1}$  \and
Daniele Evangelista$^{1}$ \and \\Marco Imperoli$^{2}$ \and Emanuele Menegatti$^{1}$ \and Alberto Pretto$^{1}$}
\institute{$^{1}$Department of Information Engineering, University of Padova, Italy,\\
\email{[fusarodani, olivastrie, evangelista, emg, alberto.pretto]@dei.unipd.it \\
$^{2}$FlexSight Srl, Padova, Italy, \email{marco.imperoli@flexsight.eu}}
}

\maketitle              % typeset the title of the contribution

\begin{abstract}
Self-driving vehicles and autonomous ground robots
require a reliable and accurate method to analyze the
traversability of the surrounding environment for safe navigation. This paper
proposes and evaluates a real-time machine learning-based
Traversability Analysis method that combines geometric features with appearance-based features in a hybrid approach
based on a SVM classifier. In particular, we show that integrating a new set of geometric and visual 
% simple but effective 
features and focusing
on important implementation details enables a noticeable boost
in performance and reliability.
%The evaluation of the proposed approach is done using a public dataset and is compared with state-of-the-art Deep Learning approaches. 
The proposed approach has been compared with state-of-the-art Deep Learning approaches on a public dataset of outdoor driving scenarios.
It reaches an accuracy of 89.2\% in scenarios of varying complexity,
demonstrating its effectiveness and robustness. The method runs fully on CPU and reaches comparable results with respect to the other methods, operates faster, and requires fewer hardware resources.
\end{abstract}
\section{Introduction}
\begin{figure*}[t]
\centering
\begin{subfigure}{.32\textwidth}
  \centering
  \includegraphics[width=\linewidth]{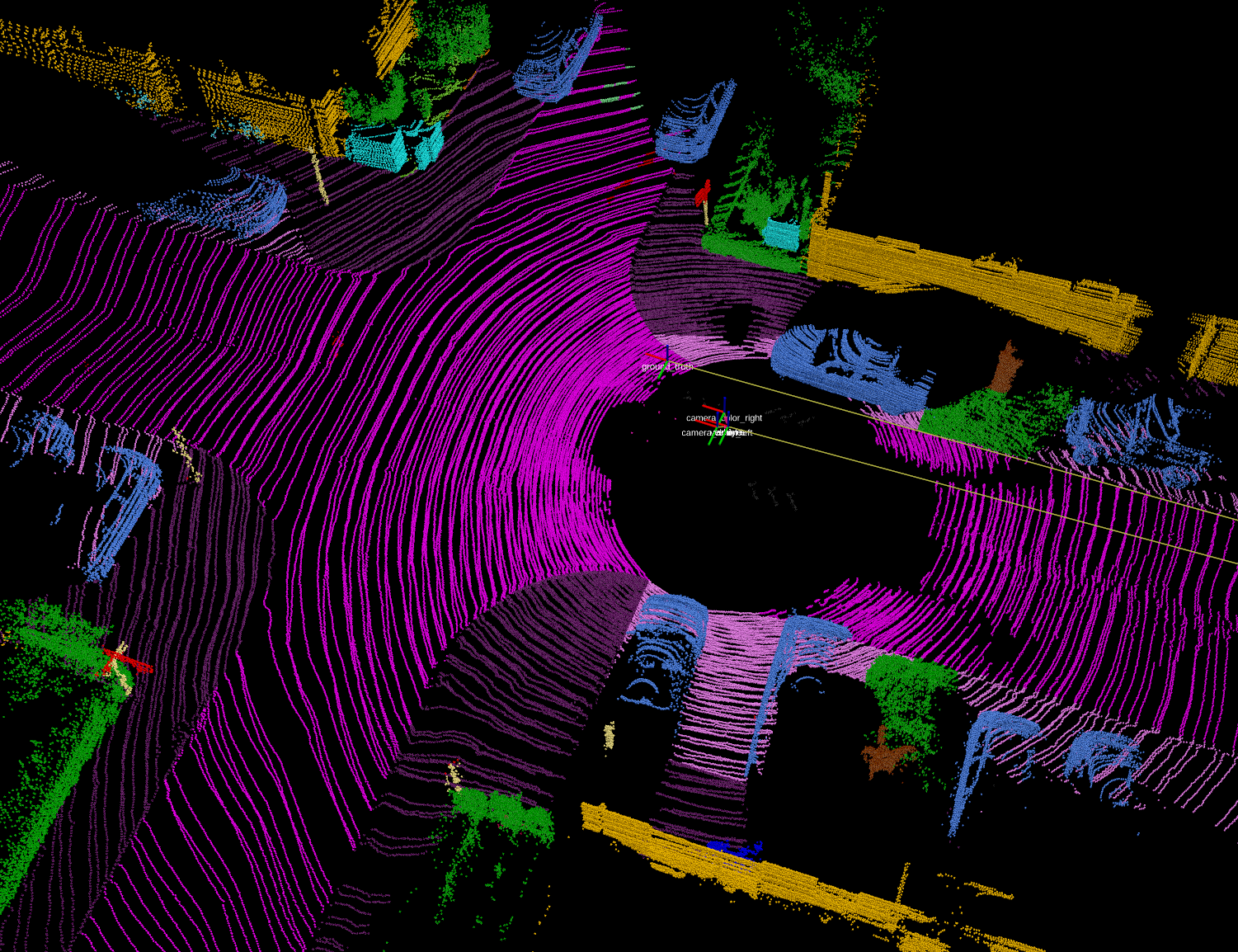}
  \caption{}
  \label{fig:kitti_labels}
\end{subfigure}
\begin{subfigure}{.32\textwidth}
  \centering
  \includegraphics[width=\linewidth]{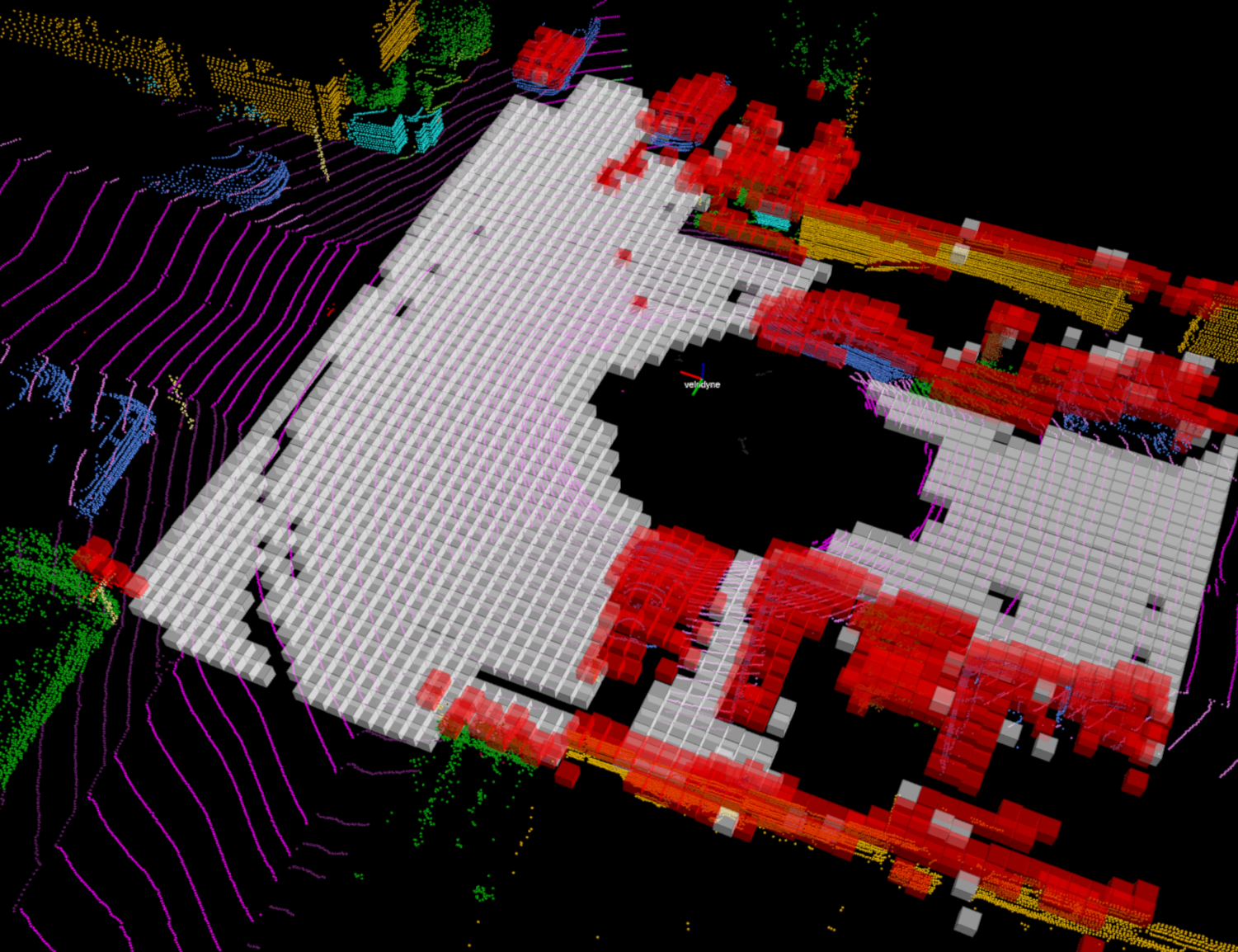}
  \caption{}
  \label{fig:gt}
\end{subfigure}
\begin{subfigure}{.32\textwidth}
  \centering
  \includegraphics[width=\linewidth]{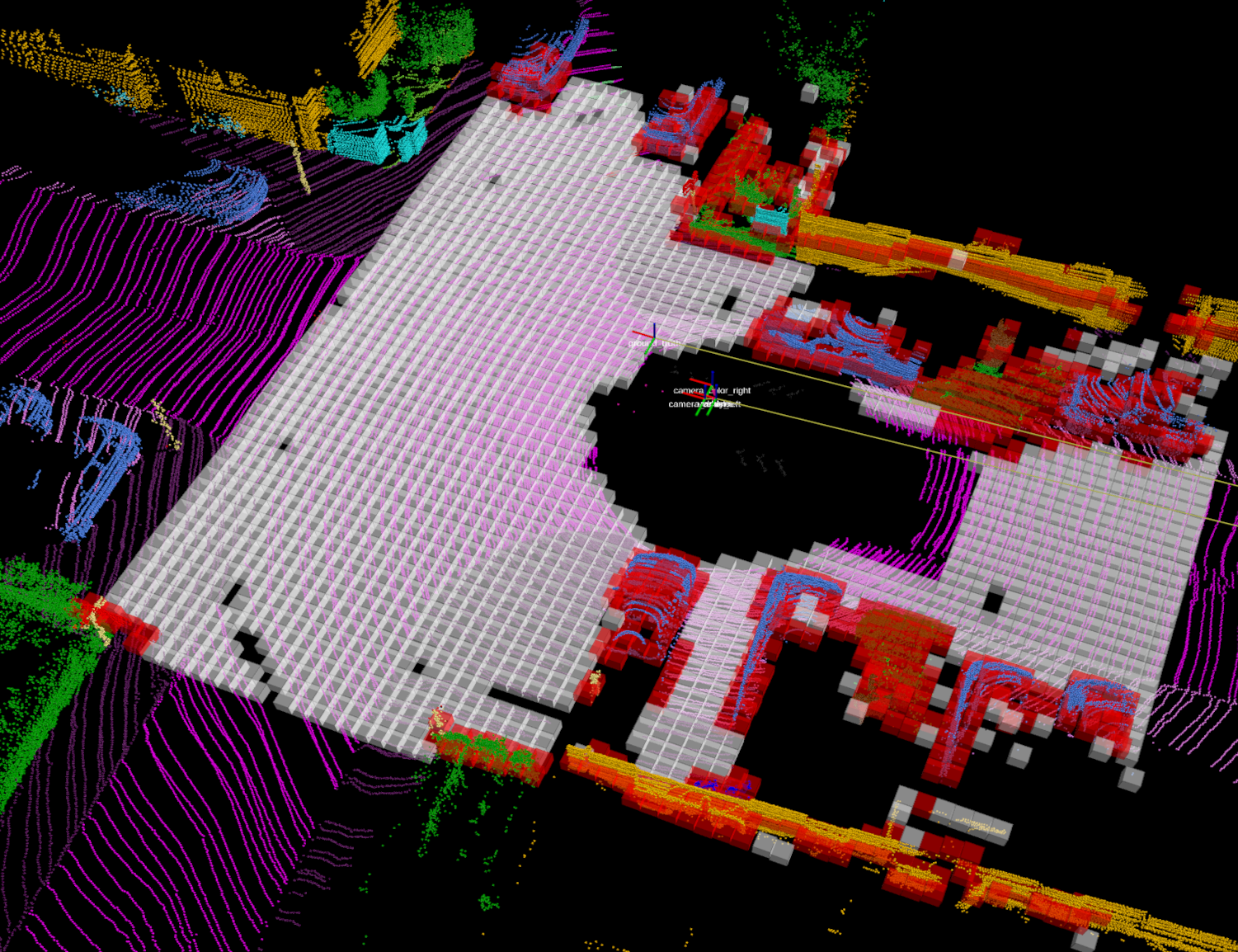}
  \caption{}
  \label{fig:mine}
\end{subfigure}
\caption{From left to right: the original SemanticKITTI Labeled cloud in a frame from Scenario 00, and the corresponding Ground Truth and predicted traversability grids respectively (white points are traversable, red ones are not)}
\label{fig:test}
\end{figure*}

Traversability Analysis is a fundamental task in the field of robotics and autonomous driving, in particular, this is one of the key activities that an autonomous vehicle has to perform to achieve an effective navigation in every type of scenario. 

The ability to correctly detect non-traversable regions of the terrain is closely related to the type of vehicle on which the analysis is performed, and, according to \cite{Papadakis2013}, its computational complexity increases with the diversity of the surrounding terrain.
As in \cite{Lee2021}, a grid cell is said to be \textit{Non Traversable} if its features exceed certain thresholds for the vehicle specification: sizes, workloads, risk awareness and so on\footnote{Note that this definition differs from the \emph{driveability} stated in \cite{drivability2003bekiaris}.}.

A binary terrain classification that discriminates between \textit{Traversable} and \textit{Non Traversable} areas needs to be reliable and efficient.
The reliability requirement is essential, it affects how much the vehicle is able to move in the environment, and it is strictly connected to the robot safety. Accordingly, the real-time requirement needs special attention as well, since this may affect the reaction speed of the vehicle.

In this paper, we propose a real-time machine-learning-based grid segmentation method, in which 3D LiDAR data and 2D camera images are used for the traversability analysis.
Our method is designed to work either with a single point cloud or with sequences of registered point clouds.
In the simplest version, a point cloud is arranged in buckets of 2D points, thus creating a 2D grid such that its reference frame coincides with the LiDAR center.
This setting is suitable for getting immediate traversability feedback, at the price of a low cloud density. 
Considering a sequence of point clouds and estimating the relative motions between them by using point cloud registration algorithms such as ICP (Iterative Closest Point) or a LiDAR odometry system such as \cite{Shan2018Loam}, it is possible to obtain a denser local map, thus a more informative 2D grid.
If a calibrated camera is available, it is possible to  re-project the point cloud into the camera image plane, so collecting also the intensity and/or color information of the points.
Our goal is to classify each cell of the collected grid: we compute a set of  features representing the group of points in each cell, then by using such features we evaluate the level of traversability of the cells using a binary classifier.
%We introduce here 4 new geometric features.  that allow outperforming other recent geometric- based approaches.
%Furthermore, by integrating a simple visual feature, it is also possible to improve even more the results, without affecting the runtime.

%The contributions of this work can be summarized as follows: (i) the introduction of four new geometric features for the training of an SVM-based traversability analysis; (ii) the development of a hybrid approach that includes two additional appearance-based features; (iii) the comparison of the proposed approach with state-of-the-art methods on a large public dataset (e.g., see Fig.  ~\ref{fig:test}); (iv) 
%the code will be released for public use\footnote{Code available at: https://bitbucket.org/flexsight/traversabilityanalysis}.

The challenge of this work is to use geometric and appearance-based features in a mixed urban environment: roads, sidewalks, mixed vegetation (e.g., trees, near-road vegetation, urban parks, forests), dynamic agents and many other urban elements that make the classification of the 3D cloud more difficult and ambiguous.

The contributions of this work can be summarized as follows: (i) the development of a hybrid approach for a SVM-based traversability analysis that includes four new geometric features along with a set of appearance-based features; (ii) the comparison of the proposed approach with state-of-the-art semantic segmentation based methods on a large public dataset; (iii) the implementation code is being released for public usage\footnote{Code available at: https://bitbucket.org/flexsight/traversabilityanalysis}.

\section{Related Work}
The concept of traversability has been proposed in the literature for measuring the general characteristics of roads \cite{Tsukada2007EVALUATIONOR}, but it has also been used for the analysis of the human driving style \cite{drivability2003bekiaris}. In the context of autonomous driving, there is no explicit formulation of traversability, some usage of this concept can be found instead in the motion planning context where "traversability maps" are used for the classification of a cell-divided map into "traversable" or "non-traversable" categories \cite{Leonard2008}, \cite{trivedi}, \cite{7823116}, \cite{thrun2006}.

Different grid-based approaches for 3D terrain traversability analysis have been proposed in the literature. A complete and comprehensive review of those methods is presented in \cite{Papadakis2013}. This survey gives a detailed overview of traversability analysis methods for autonomous vehicles within environments with highly variable complexity. In particular, the survey puts emphasis on exteroceptive sensory data processing methodologies that are presented in two main categories, namely \textit{geometry-} and \textit{appearance-based}.
%The paper also states that while geometry-based (i.e., that uses 3D LiDAR data as input) approaches provide the shape of a surface, the appearance-based (i.e., that uses 2D camera images instead) rely on better resolution data, providing a finer classification. Thus, a hybrid system should be able to accomplish both tasks.
%Moreover, the paper underlines that the predominant choice for measuring the 3D terrain traversability in road or urban contexts is producing 2D Digital Elevation Maps (DEMs), especially when 3D point clouds acquired from LiDAR data are available.

In \cite{Bellone2018}, a method for traversability analysis in urban and extra-urban scenarios is proposed. In the paper, authors evaluate the performances of various approaches based on point clouds and color data in various scenarios. They also train a SVM classifier that uses both geometric- and appearance-based local descriptors (features). Experimental results show that purely geometric-based features already provide enough accuracy, but the integration of different types of features, i.e., descriptors taken from 2D camera image processing, increases the robustness of the classifier and improves its classification accuracy. 

Another purely geometric-based approach is proposed in \cite{Chen2017}. The LiDAR point cloud is transformed into a new 2D representation called \textit{LiDAR Histogram}. In the new domain, a road line estimation can be applied to get the actual road projected into the 3D data, and the same can be done for obstacles and water hazard segmentation.

A semi-supervised learning approach is presented in \cite{Suger2015}. This approach uses a more compact feature vector and a grid representation to compute the traversability score. The classifier is trained only on positive samples, assumed to be truly traversable.

A traversability mapping method that uses Bayesian Generalized Kernel inference is found within the LeGO-LOAM framework \cite{Shan2018}. %To the best of our knowledge, this is the only method for traversability mapping whose original implementation is publicly available.
This work proposes an approach for solving the traversability analysis problem producing 2.5D elevation grid maps. It exploits Bayesian Generalized Kernel (BGK) inference to solve the sparse data problem encountered during terrain mapping (it uses only single LiDAR scans). Then, the traversability analysis is performed on those locations intersected by the current LiDAR scan points, and elsewhere it is estimated by BGK traversability inference.\\
Deep learning frameworks are one of the most used instruments to solve problems that we are not able to model. There are many Deep-Learning-based solutions for traversability analysis.

In \cite{viswanath2021offseg} images are used for Off-road traversability analysis.
Authors used a double segmentation using CNNs. The first one divides the pixels in
the image between trasversable and non trasversable, and then
the trasversable pixels are re-segments into subclasses in order to 
asses the quality of the traversability based on the type of the terrain.

In \cite{hirose2019vunet} RGB images are used for traversability map prediction. In particular
the input are the past frame, the current frame and the velocity inputs that
pass through 2 CNNs, one predicts how the static elements of the image
will change in the next frame (SNet), while the other predicts how the dynamic
elements of the scene will change (Dnet). The predicted image is fed then to a
third net that will produce the traversability map (GONet).

In\cite{hosseinpoor2021} instead of determining the traversability for specific types of robot, 
it labels which terrain is trasversable and also for which type of robot (e.g. legged, wheeled, flying ).
In \cite{paz2020probabilistic}, \cite{guan2021ttm} both images and point clouds have been fed to deep networks
in order to extract as much as information from the environment to obtain more robust traversability analysis.
In fact, in \cite{paz2020probabilistic} has been shown that this approach is effective even in unstructured environments.

In \cite{Zhu2020_DIRL} a different learning approach has been exploited. A behaviour-based method to solve traversability analysis in an offroad scenario has been proposed. This method falls in the category of Deep Inverse Reinforcement Learning because  the cost function is extracted from experts demonstrations, and then that cost function is used in order to learn a optimal policy.
In \cite{jung2021}, Deep Inverse Reinforcement Learning has been used in order to predict
the trajectories of the other agents in the scene. Using this information
to produce a traversability map that also takes into account future events. The predicted map will allow safer navigation.

%\todo{driveability metrics?}.{ Metriche importanti sono revised in [ Guo et al.] for driveability assessment.}
When developing robust autonomous driving methods, two additional elements need to be taken into account, in particular, the availability of large datasets and the usage of good metrics for testing their performance. The former is essential especially if learning-based approaches are used. An exhaustive survey of available datasets for autonomous driving benchmarking is given in \cite{8317828}, including the SemanticKITTI dataset \cite{behley2019iccv} that is used in this paper.

\section{Method}

The proposed approach leverages handcrafted geometrical and appearance-based features and Support Vector Machine (SVM) models.
In machine learning and pattern recognition, a feature is an individual measurable property or characteristic of a phenomenon \cite{BishopPR}. A geometric feature is calculated based on geometric properties, while an appearance feature is calculated starting from color or intensity properties. Handcrafted features are not new in the literature, and they may be overcome by self-learned features in the hidden layers of Deep Convolutional Neural Networks. 
However, considering comparable hardware, approaches that use these self-learned features are still slower than simple handcrafted features, and certainly not so accessible to a deep understanding of the whole process. Often the working mechanisms of the hidden layers' kernels of a Deep Neural Network are not so easily interpretable \cite{fan2021interpretability}.
Since they need large amount of data to properly generalize the self-learned decision rule, another drawback is made by the lack of rich and public labeled dataset.\\
In the following, a \textit{feature vector} of a set of points $ S = \{p_1, p_2, ..., p_n\} $ with $p_i \in \mathrm{R}^3$ is expressed in the form of a vector of features $$ \mathcal{F}(S) = [f_1, f_2, ..., f_m]^T $$where $m$ is the number of features. This vector of fixed dimension is used to train a SVM model. SVMs are popular decision models used in classification and regression problems. As stated in \cite{BishopPR}, an important property of SVMs is that the determination of the model parameters corresponds to a convex optimization problem, and so any local solution is also a global optimum. For the sake of brevity, the details on SVM models can be found in \cite{cristianini_shawe-taylor_2000}. It is sufficient to say that training a SVM model means to find an optimal decision rule to assess the label of a feature vector.
The so-called \textit{kernel trick} helps in finding such decision rule by transforming the original data into a higher dimensional space. 
By experimental results, we have found that RBF kernel was the best performing kernel with respect to Linear and Polynomial kernels. It performed good both in training accuracy and testing accuracy. For this reason, we used that one in the proposed approach.

\subsection{Grid-Based Space Representation}
The traversability of the environment is evaluated using Digital Elevation Maps (DEMs), which are a set of uniformly discretized planar grid cells on the ground: at each cell is assigned a height value, as described in \cite{Shan2018}. Here, the grid is squared and it maintains a fixed size during a simulation. Given a grid resolution $ r $ and a sensor maximum range of assessment $ maxRange $, the grid will have $ l = \frac{maxRange}{r} $ cells in each side, leading to $l^2$ cells in total as can be seen in Fig. \ref{fig:gt} and Fig \ref{fig:mine}.

\subsection{Geometric-Based Feature Extraction}
The geometric-based features $ \mathcal{F}_g(S) = [f_1, f_2, ..., f_r] $ are computed based on the geometrical properties of a set of 3D points $S$. Most of these features are calculated using the eigenvalues and the eigenvectors of the correlation matrix of the points in $ S $. This matrix is computed based on the relative space coordinates of the points in $S$, and expresses the dependency between such points.
It is a 3x3 symmetric positive semi-definite matrix with all real elements, so all of its eigenvalues are real and non-negative. Listing the eigenvalues in descending order $ \lambda_1 \geq \lambda_2 \geq \lambda_3 $, the corresponding eigenvectors $ v_1, v_2$ and $v_3$ assume a spatial significance. As explained in \cite{Bellone2018}, the eigenvector $v_1$ represents the direction of maximum variance of the points in $S$ and $v_2$ represents the direction of the second maximum variance of the points. On the contrary, $v_3$ represents the smallest direction of variance but, at the same time, assuming that the points are arranged in a smooth plane in the space, $v_3$ is also normal to that plane. In the context of terrain traversability analysis, where roads are generally planar, this assumes a certain relevance.
Thus, assuming that $\lambda_1 \neq 0$,
among others we use all the features proposed in \cite{Bellone2018}:

\begin{table}[ht]
\begin{center}
\begin{tabular}{ c c} 
    \(\displaystyle \text{linearity} = \frac{\lambda_1 - \lambda_2}{\lambda_1} \) &
    \(\displaystyle \text{sphericity} = \frac{\lambda_3}{\lambda_1} \) \\ [1.5ex]
    \(\displaystyle \text{planarity} = \frac{\lambda_2 - \lambda_3}{\lambda_1} \) &
    \(\displaystyle \text{omnivariance} = \sqrt[3]{\lambda_1 \lambda_2 \lambda_2} \) \\ [1.5ex]
    \(\displaystyle \text{anisotropy} = \frac{\lambda_1 - \lambda_3}{\lambda_1} \) &
    \(\displaystyle \text{eigenentropy} = \sum\limits_{i=1}^{3}{\lambda_i \log \lambda_i} \) \\ [1.5ex]
    \(\displaystyle \text{sum of eigenvalues} = \sum\limits_{i=1}^{3}{\lambda_i} \) &
    \(\displaystyle \text{curvature} = \frac{\lambda_3}{\sum{\lambda}}\) \\ [1.5ex]
    \(\displaystyle \text{angle} = \arccos (\vec{n} \cdot \hat{z}) \) &
    \(\displaystyle \text{goodness of fit} = \min(\sigma_i(C_q)) \) \\ [1.5ex]
    \(\displaystyle \text{roughness} = \sigma_z^2 = \frac{1}{k} \sum\limits_{i=1}{k}{(z_i - \bar{z_q}) ^ 2} \) &
    \(\displaystyle \text{normal vector} = \vec{n} \) \\ [1.5ex]
    \(\displaystyle \text{unevenness} = \frac{|r_q|}{k} \) &
    \(\displaystyle \text{surface density} = \frac{k}{d_m ^ 2 }\) \\ [1.5ex]
\end{tabular}
%\caption{List of the ground truth Traversable Labels}
%\label{table:gt}
\end{center}
\end{table}

In our approach, four new features are introduced: Zeta difference, Internal density, Curvity and Volume.\\
The \textit{Zeta Difference} feature represents the maximum difference along the elevation direction, expressed by the normal to the plane fitted to the whole point cloud. This is particularly relevant, because non traversable regions (e.g., obstacles) tend to have a wide variance along the normal direction, while points in traversable regions (e.g., roads) tend to have a very small height difference. In order to obtain a consistent metric, each point inside a cell is projected to a line, passing by the origin (the LiDAR frame origin) and having direction equal to the normal vector $v$. The projection $p_p$ of a point $p$ is obtained as follows:
$$ p_p = \frac{vv^T}{v^Tv}p \text{.}$$
Let $M$ and $m$ be the points of a cell whose projections have higher and lower values of z. Then $$z_{diff} = |M_z - m_z| \text{.}$$

The \textit{Internal density} feature expresses the discretized distribution of points internally to a cell. Let's define $r_i$ as the \textit{internal resolution} of a cell. Thus, we can subdivide the cell internally in $\frac{r}{r_i}$ cells. Then, the feature is obtained by counting how many internal cells are occupied by at least one point and dividing this by the total number of internal cells. This may be discriminant of some partially filled grid cells, or some obstacles, in which the points are very compacted.

The \textit{Curvity} feature expresses how much the LiDAR scans are arranged as smooth circular curves inside the cells. This is motivated by the observation that in traversable regions (e.g. roads) the LiDAR scan is circularly smooth, since the road is a plain surface and there are no obstacles corrupting its smoothness. In non traversable regions (e.g. grass), the scans are very chaotic, sparse, not smoothly circular curved. To express this property quantitatively, we compute a discrete histogram of the Euclidean distance of each point from the LiDAR sensor. The number of bins of the histogram is a finite number $n_b$. Then, the curvity feature is the number of empty bars. The reason behind this is that all the points in a smooth circular curve have pretty much the same Euclidean distance, and when the distances are discretized, they get counted in a single bin, leading to many empty bins. Generalizing to multiple smooth circular curves, only some bins are occupied, while many are still empty. In a chaotic situation, bins tend to be rarely left empty.

The \textit{Volume} feature expresses the volume of the smallest prism which contain all the points inside each cell. By definition, it's height is equal to the $Z_{diff}$ feature. Empirically points inside obstacle regions tend to occupy more volume than points in  traversable ones.

\subsection{Appearance-Based Feature Extraction}
Purely geometric features are often not sufficient to accurately discriminate cell classes. As shown in \cite{Bellone2018}, on average hybrid features, based on both geometry and appearance, perform better than geometric features. This fact is also confirmed by our ablation study (see Section \ref{ablation_study}).\\
We compute the appearance-based features $ \mathcal{F}_a(S) = [f_1, f_2, ..., f_r] $ based on the color properties of a set of points $S$. Unfortunately, due to the temporal desynchronization between the LiDAR and the camera and the small inaccuracies in the calibration, the 3D points are not always projected in the correct image position. 
%To mitigate this inaccuracy, we first perform point deskewing to remove the data desynchronization introduced by the spinning nature of the used LiDAR (i.e., points of the same scan are acquired at different times). The appearance-based features of each cell are then obtained by projecting a 3D prism built over the cell into the RGB image and retrieving the color information from the set of pixels within the polygonal projection of such prism (see Figure \ref{fig:color_propagation}).
To mitigate this inaccuracy, we obtain the appearance-based features of each cell by projecting a 3D prism built over the cell into the RGB image and retrieving the color information from the set of pixels within the polygonal projection of such prism (see Figure \ref{fig:color_propagation}).

Specifically, we detect for each cell the eight vertices of the prism which encloses all the points belonging to such cell. The height of the prism is by definition the value of the $Z_{diff}$ feature, and the base is obtained by looking at the distribution along x and y directions of the points. Then we project the eight vertices and check that, for each prism, at least one fits inside the image (outliers are corrected according to the image size). Finally we convert the color of each pixel within the set $S$ from the RGB color space to the Hue-Saturation-Value (HSV) color space.
%HSV is alternative, the reason is that the Hue channel encodes the important info regarding the specific color of a certain pixel, as a similarity measure of how much it is similar to one of the perceived colors. In other words, we can easily understand the color of the pixel by looking at his Hue channel.
%In this color space colors are expressed as a similarity measure of how much they are similar to the perceived colors. The Hue channel expresses the perceived color as an "independent" value with respect to luminance or saturation. Saturation instead expresses the colourfulness of a pixel related to its brightness. Value, or brightness, expresses how much a source appeares to radiate or reflect light.
The appearance-based feature used in our approach is a HSV histogram of the pixels inside the polygonal projected prism.
The histogram is defined by a number of buckets for each color channel: let $\#H$, $\#S$, $\#V$ be such quantities. The buckets $ [H_b, S_b, V_b ]$ corresponding to a pixel $ [p_H, p_S, p_V ]$ are obtained by quantizing the HSV values according to the number of buckets:

$$ [ H_b, S_b, V_b ] = \Bigg[ \floor*{\frac{p_H}{\#H}}, \floor*{\frac{p_S}{\#S}}, \floor*{\frac{p_V}{\#V}} \Bigg]$$

When all pixels have voted to the histogram, all the buckets are normalized by the number of pixels that have voted, so values are in $[0, 1]$ range.

%Considering a cell $c$, projected onto the camera image defining a polygon shape $s$ containing $n$ pixels $p \in \mathbf{N}^2$, where $p_{hue}$ is the hue value of the pixel $p$, the appearance-based features of $c$ are the Hue Average $\bar{H}$ and Variance $H_\sigma$ defined as follows:
%$$\bar{H} = \frac{\sum_{p \in s}{p_{hue}}}{n}, \quad H_\sigma = \frac{\sum_{p \in s}{(p_{hue} - \bar{H})^2}}{n}$$

\subsubsection{Color Propagation}
\begin{figure*}[t!]
\centering
  \includegraphics[width=\linewidth]{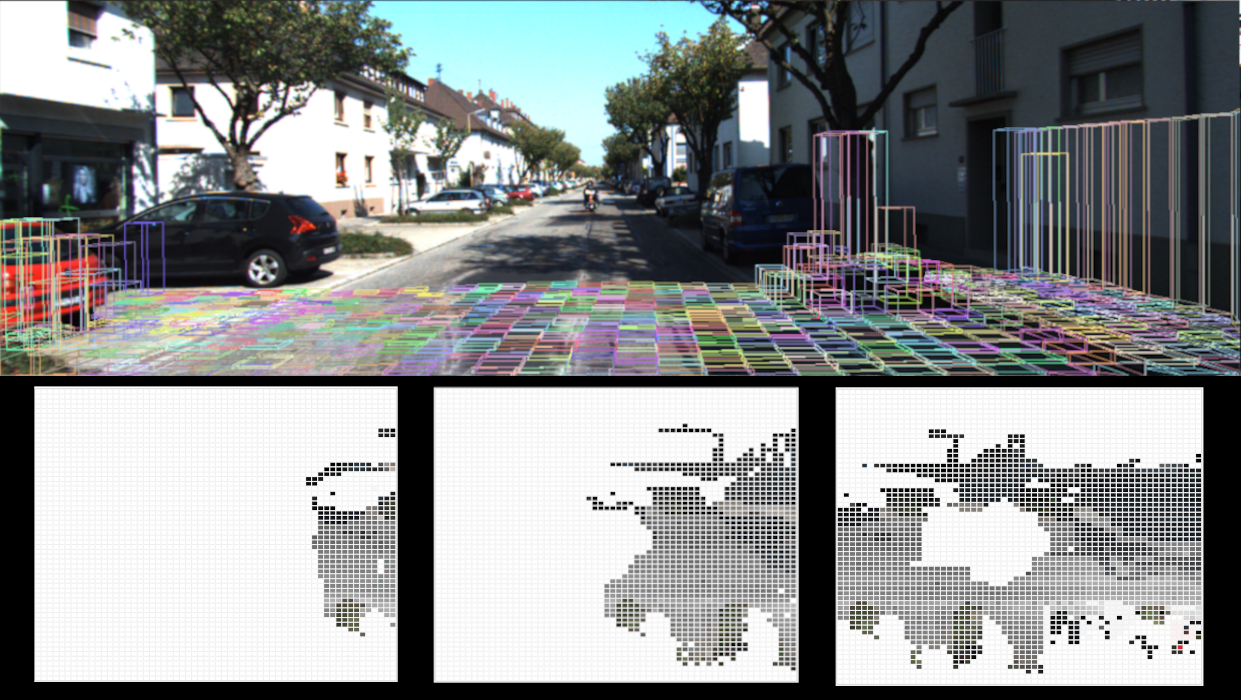}
  \caption{Color propagation at the beginning of \textit{Scenario 00}. The top image is the $28^{th}$ Left Camera image of the Scenario 00 with the 3D prisms projected (colors were chosen randomly for visualization purposes only) and the three bottom images are the colored grids at frames 03, 08 and 28, respectively (the color of each cell is obtained setting for each H, S and V channel the value of the fullest buckets). It can be seen that at the beginning few cells are colored by the intersection of grid and camera image, but as the car proceeds new cells get colored.}
  \label{fig:color_propagation}
\end{figure*}

%A single image doesn't provide enough color information: the fraction of intersected grid cells is usually around 15\% of the total.
A single image doesn't provide enough color information: the amount of grid cells that can be correctly projected to the image frame is usually around 15\% of the total cells.
For this reason a color propagation method which exploits the odometry of the vehicle helps in providing more information. Once the appearance-based features of the cells of a scan have been computed, in the following scan these features are translated to the cells which occupies their position in the current situation, and those cells are updated using the new color information (see Fig. \ref{fig:color_propagation}).
%As can be seen in Fig. \ref{fig:color_propagation}, after some frames the cells are mostly "colored".

\subsection{Point Clouds Integration}
To augment the prediction input information, $n$ subsequent point clouds are integrated forming a denser point cloud.
The $LiDAR$ reference frame, rigidly linked to the vehicle, is in motion with respect to a fixed reference frame we call  $world$. By using point set registration or LiDAR odometry algorithms, it is possible to estimate the rigid body transformations that relate a sequence of LiDAR readings. When a point cloud $P$ is obtained from the sensors, expressed in the $LiDAR$ coordinate frame, it is transformed to the $world$ frame, as it is considered a fixed reference frame. The traversability grid is also expressed in the frame $world$. 
In a stationary situation (once all $n$ point clouds have been integrated), the SVM model is able to assess the traversability of the environment. In our method we exploit the ground truth ego-motion estimation, but other methods can be used, for example LeGO-LOAM framework \cite{Shan2018Loam} as a LiDAR odometry front-end for point cloud integration.
%In our method we exploit the LeGO-LOAM framework \cite{Shan2018Loam} as a LiDAR odometry front-end for point cloud integration.

\subsection{Point Sorting into the Grid Cells}
Each point of the integrated point cloud is assigned to the cell it belongs to by looking at the point's 2D coordinates. If it lies outside the grid, it is ignored.
The row and the column of the cell to which a point belongs are calculated as follows:

$$
\begin{pmatrix} col \\ row \end{pmatrix} \approx \frac{1}{r} \begin{pmatrix} x - grid_x \\ y - grid_y \end{pmatrix}
$$

where $r$ is the map resolution, $grid_x$ and $grid_y$ the coordinates of the bottom-right cell of the grid expressed in the $world$ frame.
A cell is then considered \textit{unpredictable} if it contains fewer points than a threshold value, otherwise it is considered \textit{predictable}.

At this stage, a feature vector is computed for each \textit{predictable} cell $c$, let's call it $\mathcal{F}_c$. A trained SVM model can predict the traversability value of the cell $c$  using $\mathcal{F}_c$ as input.

\subsection{Outliers Filter}
Given the difficulty of the task, the SVM model can sometimes produce some outliers. For example, in some isolated cells of a plain traversable road region, it can predict a non traversable value. For this reason, a simple yet really effective signaling-based post processing filter is applied.
Here we make the assumption of ignoring very small obstacles ( size $\leq $ resolution $ r $ ).
Once all the cells have been assigned a traversability value by the SVM model, let us call it $t^{0}_{ij}$ with $i, j \in [0, ..., l] $, each cell is assigned a new value, let it be $t^{1}_{ij}$. This is computed considering the predicted label of the cell $t^{0}_{ij}$ weighted with a factor $w$ and the labels of the 8-neighbors of the cell $c_{ij}$ weighted with a factor $1$. $t^{1}_{ij}$ is then the most frequent label.

Consider a cell $t^{0}_{ij}$ labeled as \textit{non traversable} ($\bar{T}$) surrounded by all \textit{traversable} ($T$) cells. Then the label-distribution for such cell would be, using $w = 3$, \{$\bar{T}$, $\bar{T}$, $\bar{T}$, $T$, $T$, $T$, $T$, $T$, $T$, $T$, $T$\}. Then, having 3 $\bar{T}$ and 8 $T$, we will assign $t^{1}_{ij} = T$, that is, the cell will be labeled as \textit{traversable}, even if the SVM-model labeled it as \textit{non traversable}.

\section{Experiments}

To evaluate the performance of the proposed approach, the SemanticKITTI dataset is used \cite{behley2019iccv}. All the experiments of this method are tested on an AMD Ryzen \textsuperscript{\texttrademark} 7 5800H, $ 8 * 3.2 GHz$, 16GB RAM platform. All the Deep Learning based approaches are tested on an Intel\textsuperscript{\textregistered} Core\textsuperscript{\texttrademark}) i9-10920X CPU (3.50GHz) with Nvidia Titan RTX. All algorithms are implemented in ROS Noetic, in C++, in a Linux environment, with no GPU support and no CPU parallelization.

\subsection{SemanticKITTI Dataset}
The dataset used to evaluate the approach proposed is 
%the
SemanticKITTI, a public dataset for Semantic Scene Understanding using LiDAR Sequences. This dataset is based on the KITTI Odometry Benchmark \cite{kitti_odometry}.
It comes up with 11 different scenarios, named \textit{scenario00-scenario10}, in which a vehicle with the sensor setup installed on board is driven within an urban context, in low traffic conditions. There are sometimes dynamic agents (other vehicles moving around, people, etc.) and natural elements (grass, parks, trees, etc.).
Among the info provided by the dataset, the method makes use of:
\begin{itemize}
    \item LiDAR clouds, where each point carries the 3D Euclidean coordinates expressed in the LiDAR frame along with the RGB value of the class it belongs to;
    \item Left RGB Camera Images;
    \item Calibration Matrix, expressing the relative pose between the left RGB camera frame and the LiDAR frame
    %\item Traversability Ground Truth ego-motion estimation.
    \item Ground Truth ego-motion estimation.
\end{itemize}

\subsection{Traversability Ground Truth Extraction} \label{subsection:gt}
The points in the SemanticKITTI dataset are not labeled with the \textit{traversable} and \textit{not traversable} labels. Indeed, the definition of traversability is in itself quite ambiguous. In order to obtain a ground truth dataset to evaluate the method, a pre-processing of the labels was needed. We considered as \textit{traversable} the points having the label corresponding to one of the following classes of the SemanticKITTI dataset: \textit{road, sidewalk, parking, lane marking, other ground}. %in Table \ref{table:gt} (please refer to \cite{behley2019iccv} for a complete list of all the available labels):
Since our method refers to grid-based traversability analysis, we need to generalize the concept at a grid cell level.
%The label of a cell is calculated by looking at the most elevated point in the z coordinate (in the fixed \textit{world} reference frame). If it is a \textit{traversable} point, then the cell is considered \textit{traversable}, otherwise it is considered \textit{non traversable}. This is done because the dataset is not perfect, there are some mistakes: some single points are labeled as obstacles while they are perfectly traversable, or there are some LiDAR outliers.
%Please note that while the proposed approach can assess the traversability of a cell only when a minimum number of points is found inside of it, the ground truth reports the label of every cell having at least one point. All the empty cells are considered \textit{unpredictable}. This fact makes the task even more challenging.

The label of a cell is calculated by looking at the LiDAR points that fits inside each squared cell. Let $m$ be the minimum number of points for a cell to be considered predictable. In our approach we set $m=2$. If a cell contains less than $m$ points, it is labeled as \textit{unpredictable}. Otherwise, if the cell contains at least 2 \textit{non traversable} points, the cell is labeled as \textit{non traversable}. Otherwise, the cell is labeled \textit{traversable}.
This is done because some outliers are collected in the SemanticKitti Dataset: sometimes a fully \textit{traversable} cell (e.g., road cells) contains an outlier marked as \textit{non traversable}. By filtering out based on at least 2 \textit{non traversable} points, we get rid of such outliers.
The Ground Truth collected in this way comes with drawbacks and benefits. Consequences include:
\begin{itemize}
    \item There may be points labeled as \textit{traversable} that are actually under a car.
    %points that belongs to a road under a car and are all \textit{traversable} label the cell as \textit{traversable}, while it may be just in the middle of a car. This is unsolvable, unless we proceed by manually segment each grid (consider every possible situations)
    % \item road/sidewalk transient is not considered. 
    \item A number of $m = 2$ points is sometimes not sufficient for some geometric properties to completely discriminate cells.
    This can be solved by using appearance-based features and a maxRange relatively close to the vehicle. If the car proceeds at high speeds, the sparsity of the cloud may be insufficient. In our approach this is solved by clouds integration and training using a bigger maxRange than the one used during test evaluation.
    \item Semantic-based methods produce a segmented cloud as output. Such methods produce a label for each point, so in order to compare our approach with their results we need to transfer the labels in the same way as we did for the SemanticKitti Ground Truth.
    \item Ground Truth for integrated clouds involve the formation of phantoms of points (e.g., when a car moves fast, the resulting integrated cloud maintains the points in all the subsequent frames). In order to be consistent with the different configurations, we decided to build the Ground Truth upon integrated or single cloud based on the need of each system. It is advisable that in the future a phantom-recognizer system should get rid of such outliers.
\end{itemize}
At this stage, each feature vector used in the previous sections can now be assigned a ground truth traversability label.

\subsection{Training and Parameters}
The experiments have been carried out using scripts that (a) automatically extract features for training, (b) train the model and (c) test the performances on unknown data.
Each scenario of the dataset is used in the first 5 seconds to produce some amount of features, to obtain a dataset to train a SVM model.
The training is carried out in a grid search approach. The fixed parameters are listed in Table \ref{table:fixedparams}.

\begin{table}[ht]
\begin{center}
\begin{tabular}{|| l | c ||} 
 \hline
 grid length & $ maxRange = 12m $ \\ [0.5ex] 
 \hline
 map resolution & $ r = 0.4m$ \\ [0.5ex] 
 \hline
 internal cell resolution & $ r_i = 0.2m$ \\ [0.5ex] 
 \hline
 min points in a predictable cell & $ min = 2$ \\ [0.5ex] 
 \hline
 point clouds integrated & $ n = 3 $ \\ [0.5ex] 
 \hline
 curvity histogram bars & $ n_b = 160 $ \\ [0.5ex] 
 \hline
 %k-folds for model validation & $ k = 5 $ \\ [0.5ex] 
 HSV histogram buckets & \#H=32, \#S=8, \#V=48 \\ [0.5ex]
 \hline
\end{tabular}
\caption{List of the used parameters}
\label{table:fixedparams}
\end{center}
\end{table}

Using a grid based approach in a 5-folds validation setup, the best parameters for each model were selected. The best model was then retrained using the whole training set.
Since the first 5 seconds of each scenario are used for the training phase, the performance evaluation is done using the following 50 seconds of the sequences.

\subsection{Metrics}
To compute the metrics we used the number of True Positives (TP), True Negatives (TN), False Positives (FP), False Negatives (FN) and Unknowns (UNK). The latest are those having a label in the ground truth but unpredictable by the method. Let $TOT = TP + TN + FP + FN + UNK$, the performance is measured using the following metrics:
\begin{itemize}
    \item Accuracy (Acc): 
    %cells having no ground truth labels (e.g., that do not contain any point) are ignored in the accuracy evaluation. A cell labeled \textit{unpredictable} by the method, for which there exists a \textit{known} label in the ground truth, is considered wrong. %All the other cells are parsed as usual % ;
    $$ Acc = \frac{TP+TN}{TOT} $$
    \item Intersection over Union (IoU): 
     $$ IoU = \frac{TP}{FP + FN + TP} $$
    \item F1 score: (harmonic mean of precision and recall) $$ F1 = \frac{2TP}{2TP + FP + FN} $$
    \item Rates over the total number of samples: $$(TPR, TNR, FPR, FNR) = \frac{(TP, TN, FP, FN)}{TOT}$$
    \item Latency of the system inferencing the labels: it refers to the time taken to process one unit of data provided that only one unit of data is processed at a time. The metric is calculated by averaging the inference times of each label. Recall that the LiDAR data rate is approximately 10Hz, thus the maximum time allowed for inferencing a cloud is 100ms. 
\end{itemize}

\subsection{Comparison}
\label{sec:experiments}
The experiments carried out aim to emphasize the performances brought by the contributions of this work, in particular the use of the 4 new geometric-based features and the appearance-based features, the post-processing filter and the low latency due to the SVM models. 

We tested our framework against different recent methods: RangeNet++ \cite{Milioto2019RangeNetF} (comparing many backbones) and PointNet \cite{qi2017pointnet}.% and Cylinder3D \cite{zhu2020cylindrical}.

RangeNet++ is a segmentation-based network: it outputs the label of each point of the cloud that was given as input.
Every RangeNet++ model was trained from scratch using as learning labels only two classes (traversable and not traversable) by remapping the original labels as explained in section \cite{Milioto2019RangeNetF}. RangeNet++ has been trained using different backbones (DarkNet and Squeezeseg, in different versions).
To compare them with the proposed approach we treat the RangeNet++ predicted labels as if they were the ground truth, assigning labels to cells as described in section \ref{subsection:gt}.
PointNet, instead, is a classification network. When given the set of points belonging to each cell the network predicts the corresponding label. In order to do so, we need to create a new dataset from SemanticKitti where points were packed up to the cell they belong to, together with their ground truth label. A fixed-length vector of points is fed to the network (by trial and errors, we found that the best accuracy is reached with a fixed length of 256) and the result of the network is the label of each cell.

\begin{figure*}[t]
\centering
\begin{subfigure}{.45\textwidth}
  \centering
  \includegraphics[width=\linewidth]{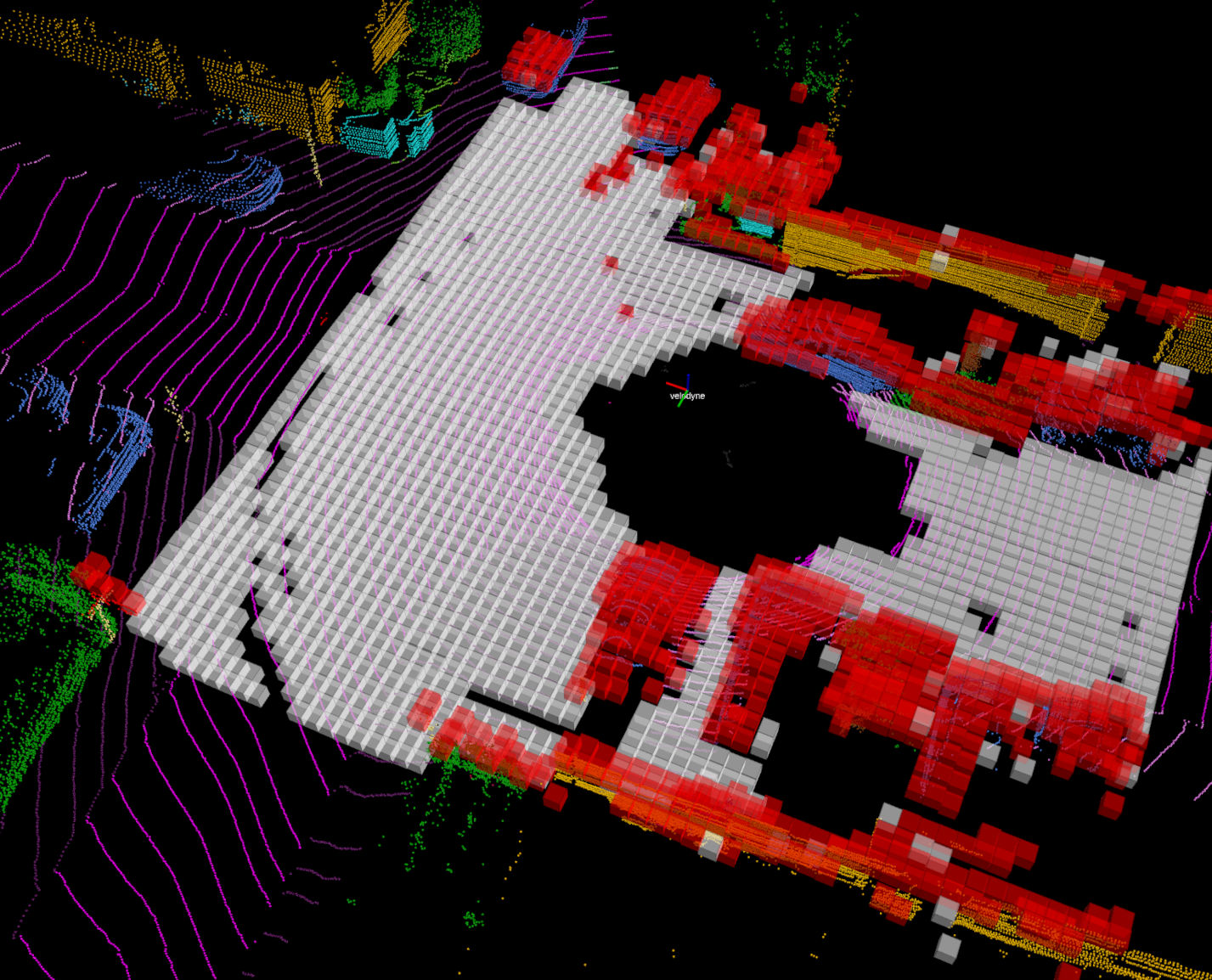}
  \caption{}
  \label{fig:ground-truth-grid}
\end{subfigure}
\begin{subfigure}{.45\textwidth}
  \centering
  \includegraphics[width=\linewidth]{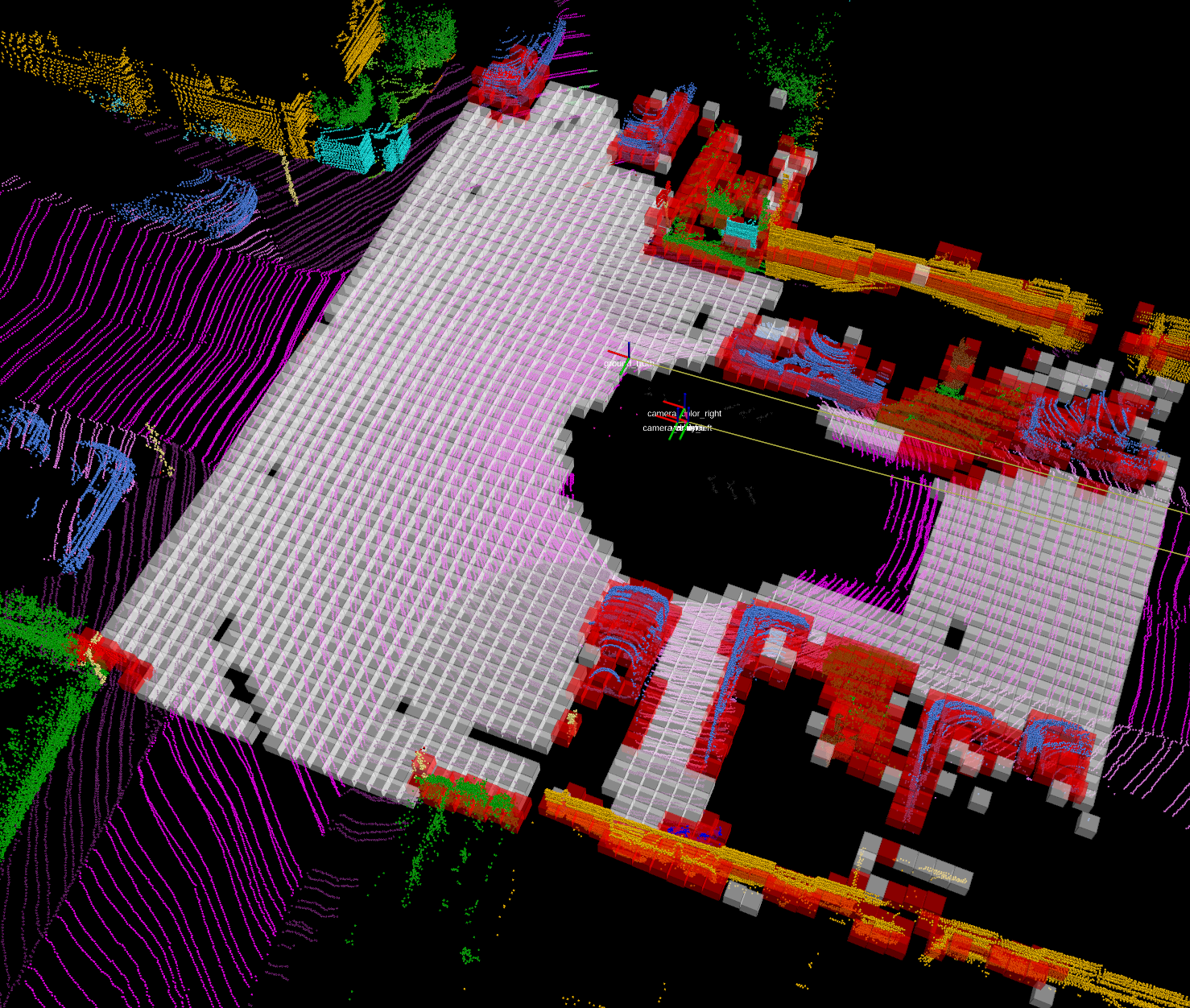}
  \caption{}
  \label{fig:mine1}
\end{subfigure}
\begin{subfigure}{.45\textwidth}
  \centering
  \includegraphics[width=\linewidth]{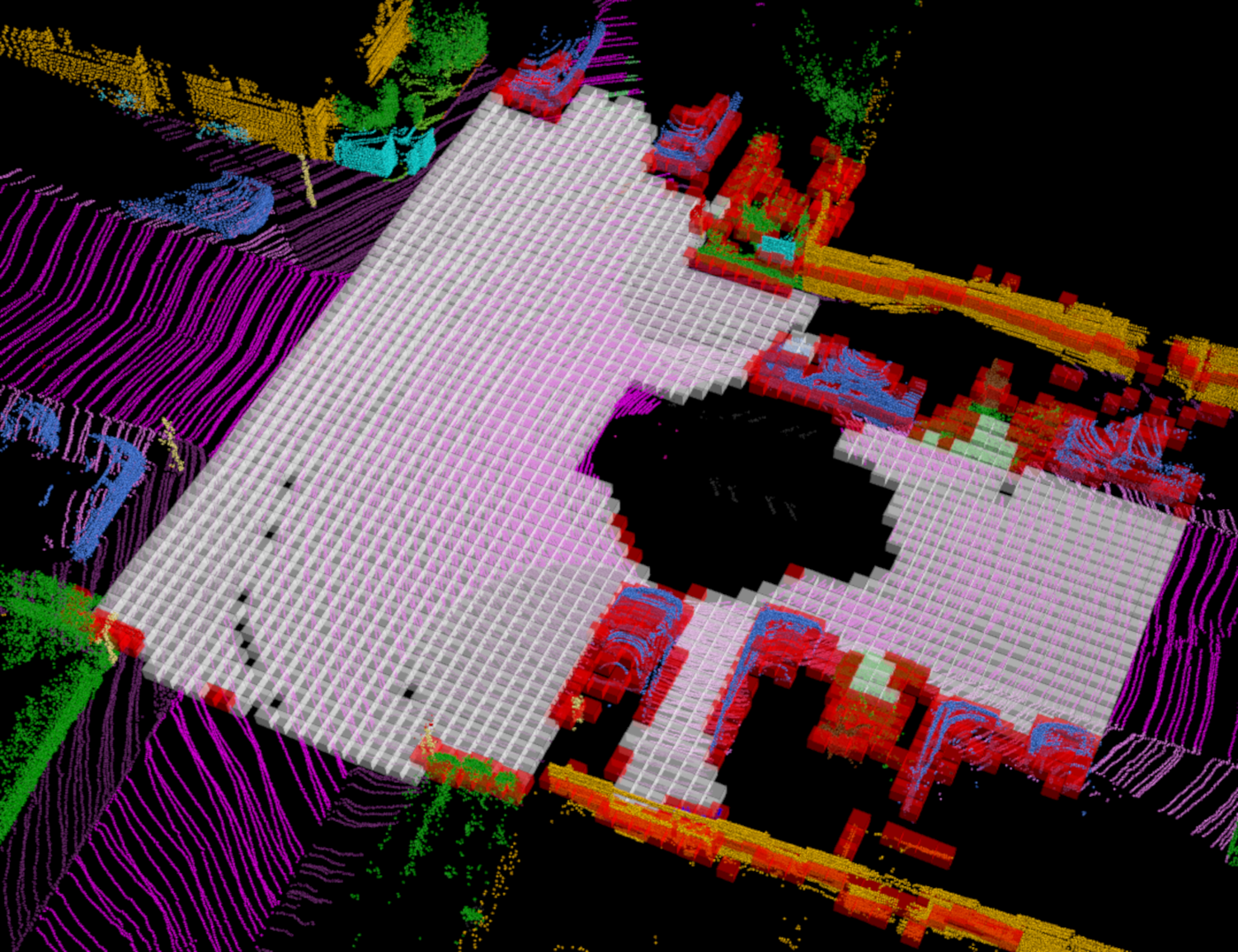}
  \caption{}
  \label{fig:pointnetres}
\end{subfigure}
\begin{subfigure}{.45\textwidth}
  \centering
  \includegraphics[width=\linewidth]{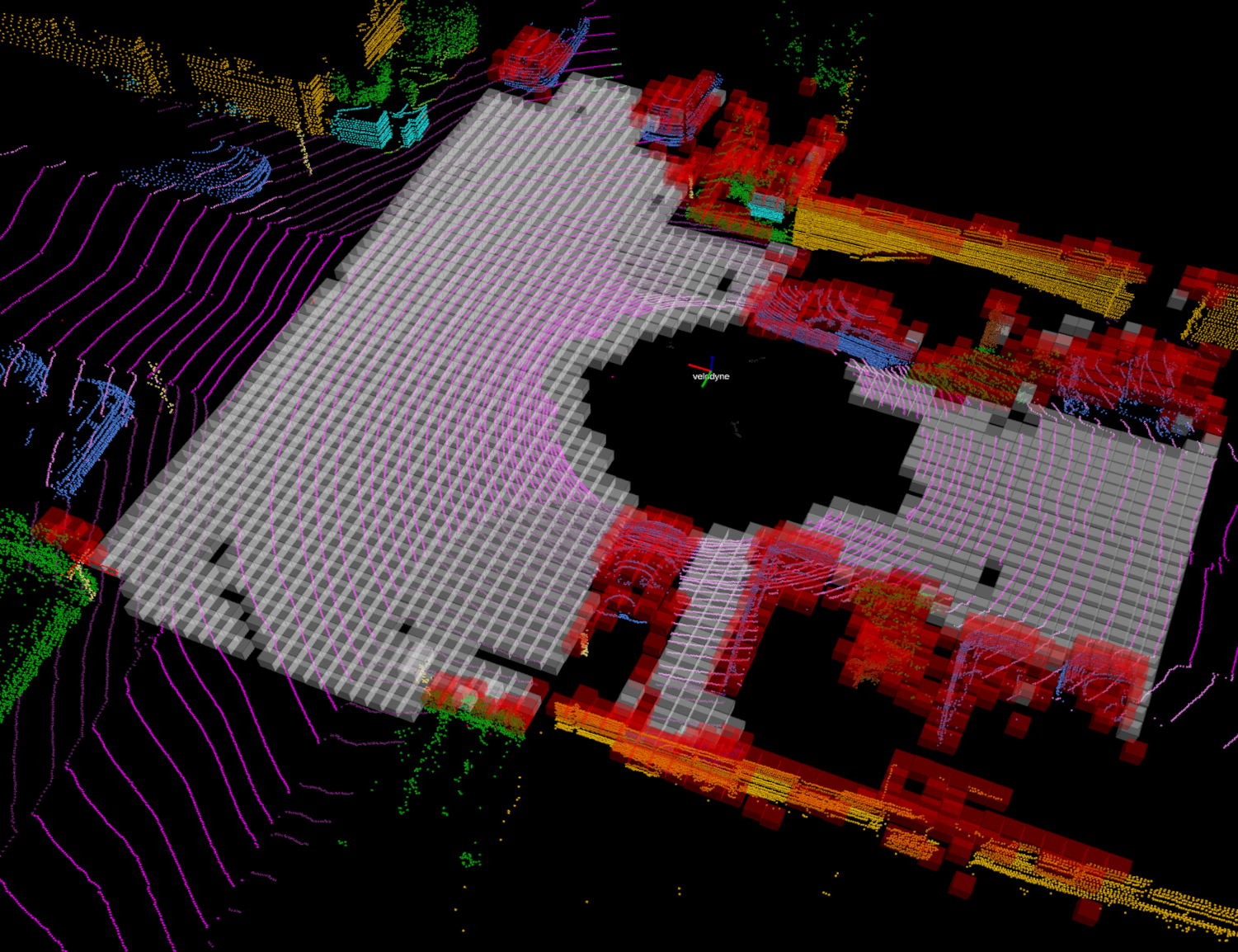}
  \caption{}
  \label{fig:rangenet-predicted-cloud}
\end{subfigure}

\caption{Figure \ref{fig:ground-truth-grid} depicts an example of Ground Truth Traversability (white points are traversable, red ones are not), Figure \ref{fig:mine1} is the result of the method proposed, Figure \ref{fig:pointnetres} the result of the PointNet, Figure \ref{fig:rangenet-predicted-cloud} the result of RangeNet++ (with DarkNet53-1024px backbone).}
\label{fig:test1}
\end{figure*}

Table \ref{table:results} shows that the results of the proposed method is comparable with the other methods results. Deep Learning based method have slightly better accuracy, but it comes at the cost of a bigger latency and expensive required hardware.
SVM latency is around 19 ms, while the other methods' latency vary across 20-105 ms for RangeNet++ and 2000ms for PointNet.

The overhead of the system is very important in this context. If the LiDAR and the camera provide new information with a rate of 10Hz, only 100ms are available for the traversability computation. But 100ms is even too much time: very often an Autonomous Driving system has to compute many other things, and so the available time is a smaller portion of the total available. Not to mention the time needed for the system to contextualize the results: for path planning and other tasks occupancy grids like the one proposed are used, and so the system needs also to compute such information from the pure segmentation cloud.

It is clear how the best accuracy is reached by RangeNet++ using the Darknet53-1024px backbone. It's beaten only by the DarkNet53 in TNR and FPR rates. Both are slower compared to our method.
If the same duration metric was meant to be obtained, the accuracy of RangeNet++ would be very similar to our method, while the mIoU drops lower our mIoU.
By comparing \cite{Milioto2019RangeNetF}'s results and ours, we identify the same so-called \textit{label re-projection problem}. Thus the Ground Truth extraction may penalize this drawback of the method. The same extraction affects out method: when planar \textit{non traversable} regions are met illumination drastically changes.

PointNet is rather slow and not easily exploitable in a real context, although it performs very well from an accuracy point of view.

The proposed approach is feasible: the latency is very low and the accuracy is competitive. The whole process of clouds integration, feature extraction, SVM prediction and grid development takes around 60ms, perfectly suitable for real-time real-world application. The hardware needed is a CPU-only PC, a 3D LiDAR scan and an RGB Camera.

\begin{table}[t]
\begin{center}
\begin{tabular}{||C{2.8cm}|C{0.8cm}|C{1cm}|C{0.8cm}|C{1cm}|C{1cm}|C{1cm}|C{1cm}|C{1.5cm}   ||}
\hline
\textbf{Experiment} & \textbf{Acc} & \textbf{mIoU} & \textbf{F1} & \textbf{FPR} & \textbf{TPR} & \textbf{FNR} & \textbf{TNR} & \textbf{Latency} \\ [0.5ex]
\hline
%trav\_analysis & 88,1 & 83,0 & 90,2 & 10,8 & 92,6 & 7,4 & 89,2 & \textbf {19 ms} \\ [0.5ex]
trav\_analysis & 89,2 & 84,9 & 91,4 & 10,2 & 93,6 & 6,4 & 89,8 & \textbf{19 ms} \\ [0.5ex]
\hline
Darknet21 & 92,7 & 86,0 & 92,2 & \textbf{3,1} & 88,3 & 11,7 & \textbf{96,9} & 40 ms \\ [0.5ex]
Darknet53 & \textbf{93,4} & \textbf{87,4} & \textbf{93,0} & 3,7 & 90,2 & 9,8 & 96,3 & 105 ms \\ [0.5ex]
Darknet53-512px & 91,6 & 83,8 & 90,9 & 3,8 & 86,9 & 13,1 & 96,2 & 25 ms \\ [0.5ex]
Darknet53-1024px & 93,1 & 86,9 & 92,7 & 3,4 & 89,3 & 10,7 & 96,6 & 55 ms \\ [0.5ex]
Squeezeseg & 90,1 & 81,4 & 89,4 & 5,2 & 85,3 & 14,7 & 94,8 & 20 ms \\ [0.5ex]
SqueezesegV2 & 92,3 & 85,5 & 91,9 & 4,3 & 88,1 & 11,9 & 95,7 & 25 ms \\ [0.5ex]
\hline
%Cylinder3D & xxxx & xxxx & xxxx & xxxx & xxxx & xxxx & xxxx & 275 ms \\ [0.5ex]
%\hline
PointNet & 90,0 & 87,4 & 93,0 & 7,3 & 93,8 & 6,3 & 92,7 & 2000 ms \\ [0.5ex]
\hline
\end{tabular}
\caption{Average results on SemanticKITTI among scenarios 00 - 10 (all values are percentages)}
\label{table:results}
\end{center}
\end{table}

\subsection{Ablation Study} \label{ablation_study}
\begin{table}[t]
\begin{center}
\begin{tabular}{||C{2.8cm}|C{0.8cm}|C{1cm}|C{0.8cm}|C{1cm}|C{1cm}|C{1cm}|C{1cm}|C{1.5cm}   ||}
\hline
\textbf{Experiment} & \textbf{Acc} & \textbf{mIoU} & \textbf{F1} & \textbf{FPR} & \textbf{TPR} & \textbf{FNR} & \textbf{TNR} & \textbf{Latency} \\ [0.5ex]
\hline
trav\_analysis & 89,2 & 84,9 & 91,4 & 10,2 & 93,6 & 6,4 & 89,8 & 19 ms \\ [0.5ex]
\hline
only geom & 85,8 & 79,5 & 87,9 & 11,8 & 89,4 & 10,6 & 88,2 & 13 ms \\ [0.5ex]
\hline
\end{tabular}
\caption{Average results on SemanticKITTI among scenarios 00 - 10 (all values are percentages) - ablation study to demonstrate the effectiveness of the appearance-based features}
\label{table:results_ablation_study}
\end{center}
\end{table}

In order to demonstrate the effectiveness of the appearance-based features in a hybrid approach in contrast to a purely geometric-based method, we have conducted an ablation study. As shown in Table \ref{table:results_ablation_study}, the purely geometric method performs worse than the hybrid one. This happens because several \textit{Not Traversable} flat, equally distributed cells, e.g. containing pretty flat grass points, cannot be distinguished from \textit{Traversable} road cells using only geometric information.
%: probably they are both flat, equally distributed, and so on. 

\section{Conclusions}

In this paper, a real-time Machine Learning Traversability Analysis method has been proposed, which combines geometric and appearance based features. The method has been evaluated on a public dataset and the performances have been compared with state-of-the-art methods. The proposed method is comparable with the others in terms of accuracy metrics,
while it is faster and cheaper in terms of hardware resources since it runs completely on CPU without the need for high-end GPU like other methods. The tests have shown the effectiveness of the newly introduced features, the exploited grid-based data integration method and the simplified framework make the overall approach easy to integrate within real autonomous driving pipelines with minimal effort but with competitive results in terms of classification of traversable areas in the map.
Possible improvements can be obtained with the integration of self-learning subsystems, in particular for image-based scene understanding, to better exploit the color information when available.

\section*{Acknowledgments}
This project has received support from the FlexSight Srl company.
Part of this work has been supported by MIUR (Italian Minister for Education) under the initiative “Departments of Excellence” (Law 232/2016).

\bibliography{references}
\bibliographystyle{IEEEtran}

\end{document}